%% file: egpaper_for_review.tex
\documentclass[10pt,twocolumn,letterpaper]{article}

\usepackage{wacv}
\usepackage{times}
\usepackage{epsfig}
\usepackage{graphicx}
\usepackage{amsmath}
\usepackage{amssymb}
\usepackage{subcaption}
\usepackage{caption}
\usepackage{nidanfloat}\usepackage[bottom]{footmisc}
\usepackage{algorithm}
\usepackage{algorithmic}

\usepackage[pagebackref=true,breaklinks=true,letterpaper=true,colorlinks,bookmarks=false]{hyperref}

\newcommand*\samethanks[1][\value{footnote}]{\footnotemark[#1]}

\wacvfinalcopy 

\def\wacvPaperID{318} 

\ifwacvfinal\fi
\begin{document}

\title{DeepErase: Weakly Supervised Ink Artifact Removal in Document Text Images}

\author{W. Ronny Huang\thanks{Authors contributed equally.} \hspace{1cm} Yike Qi\samethanks \hspace{1cm} Qianqian Li \hspace{1cm} Jonathan L. DeGange \\
Ernst \& Young LLP \\
{\tt\small \{ronny.huang, yike.qi, qianqian.li, jonathan.degange\}@ey.com}
}

\maketitle
\ifwacvfinal\thispagestyle{empty}\fi

\begin{abstract}
Paper-intensive industries like insurance, law, and government have long leveraged optical character recognition (OCR) to automatically transcribe hordes of scanned documents into text strings for downstream processing. Even in 2019, there are still many scanned documents and mail that come into businesses in non-digital format. Text to be extracted from real world documents is often nestled inside rich formatting, such as tabular structures or forms with fill-in-the-blank boxes or underlines whose ink often touches or even strikes through the ink of the text itself. Further, the text region could have random ink smudges or spurious strokes. Such ink artifacts can severely interfere with the performance of recognition algorithms or other downstream processing tasks. In this work, we propose DeepErase, a neural-based preprocessor to erase ink artifacts from text images. We devise a method to programmatically assemble real text images and real artifacts into realistic-looking ``dirty'' text images, and use them to train an artifact segmentation network in a weakly supervised manner, since pixel-level annotations are automatically obtained during the assembly process. In addition to high segmentation accuracy, we show that our cleansed images achieve a significant boost in recognition accuracy by popular OCR software such as Tesseract 4.0. Finally, we test DeepErase on out-of-distribution datasets (NIST SDB) of scanned IRS tax return forms and achieve double-digit improvements in accuracy. All experiments are performed on both printed and handwritten text.

\end{abstract}

\begin{figure*}[hb!]
    \begin{center}
      \includegraphics[width=0.69\linewidth]{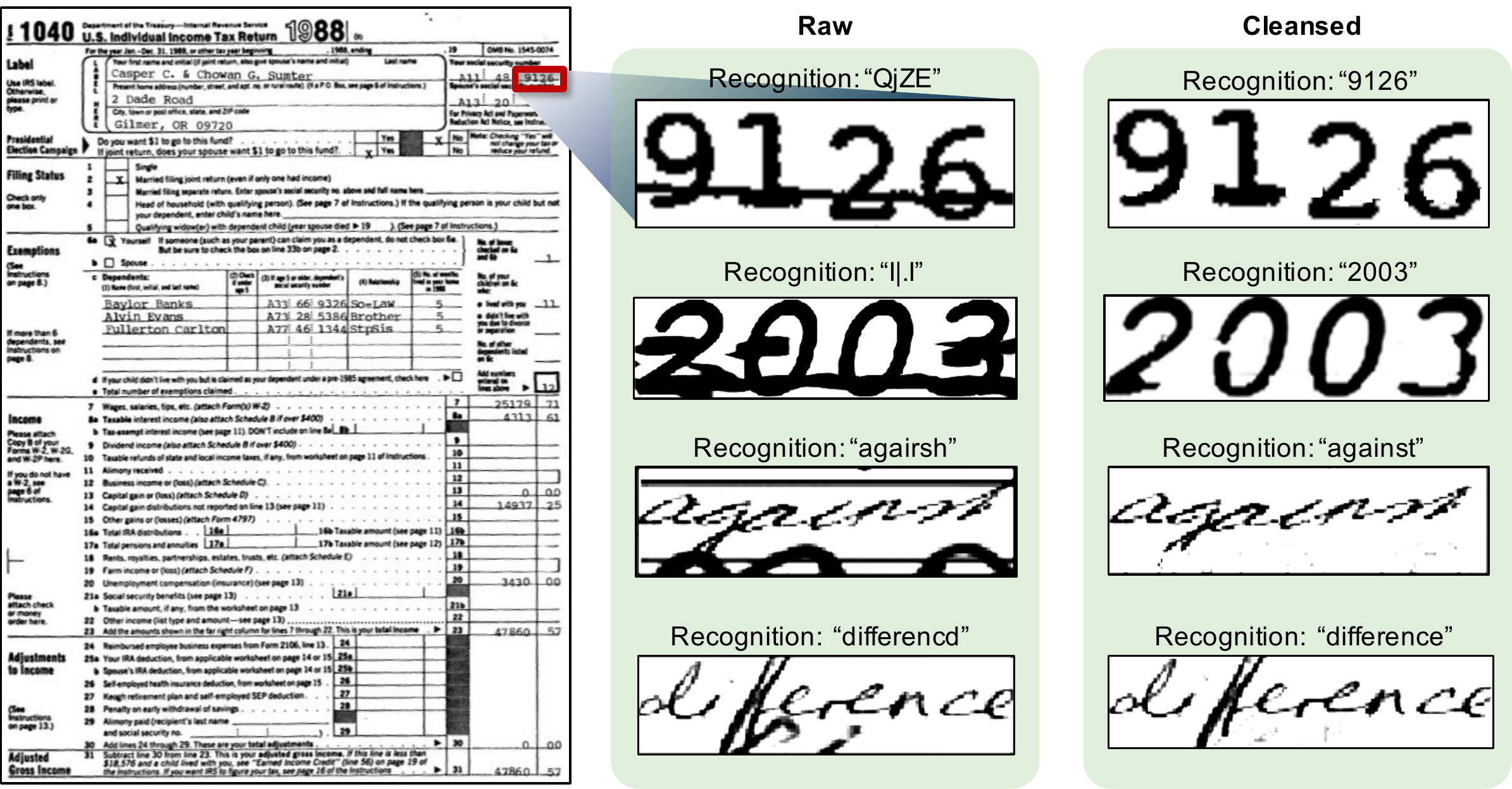}
      \vspace{-10pt}
    \end{center}
   \caption{DeepErase cleans, or \textit{erases}, ink artifacts from document text images, improving recognition accuracy, visual appeal, and other downstream tasks. Here we show text images cropped from scanned documents with various ink artifacts, such as underlines, boxes, smudges, and spurious strokes. DeepErase removes those artifacts, immediately improving recognition performance by Tesseract 4.0, a widely used open-source OCR tool, for printed text and by \texttt{SimpleHTR}, a popular offline handwriting classifier, for handwritten text.}
    \label{fig:cover}
\end{figure*}

\vspace{-15pt}

\section{Introduction}
Despite the digitization of information over the past twenty years, large swaths of industry still rely on paper documents for data entry and ingestion. Optical character recognition (OCR) has thus become a widely adopted tool for automatically transcribing text images to text strings. Modern convolutional neural networks have driven many major advances in the performance of OCR systems, culminating in the large-scale adoption of OCR tools such as Tesseract 4.0, Abbyy Fine Reader, or Microsoft Computer Vision OCR.

OCR, or more generally document text recognition, relies on a two-step process: (1) Localization: determine regions of the image (i.e. bounding boxes) which contain text and crop out those regions. (2) Recognition: transcribe cropped text image into a text string. Localization was traditionally performed via sliding window-based techniques and nowadays is performed via region proposal networks \cite{tian2016detecting}. Meanwhile, convolutional feature extractors \cite{lecun1990handwritten} coupled with recurrent classifiers with the connected temporal classification (CTC) loss has long been the workhorse of text recognition algorithms \cite{graves2006connectionist, graves2009offline}, although more recent approaches use attention-based networks \cite{bluche2017scan, poulos2017attention}.

The relevant text to be extracted from real world documents are often nestled inside of rich formatting such as tabular structures or forms with fill-in-the-blank boxes or underlines. Furthermore, documents with handwriting entries often contain handwritten strokes which do not stay within confines of the boxes or lines in which they belong and can encroach into regions occupied by other text that needs to be transcribed (henceforth such encroachment strokes will be called \textit{spurious strokes}). When extracting text regions from such richly formatted documents, it is inevitable that such document \textit{ink artifacts} are present in the cropped image even if the localization is perfect. Such artifacts can severely degrade the performance of recognition algorithms, as shown in Figure \ref{fig:cover}.

Despite the prevalence of these artifacts in the real world, many document text recognition datasets, including IAM \cite{marti2002iam}, NIST SDB19 \cite{johnson2012nist}, and IFN/ENIT \cite{abed2007ifn} contain only images which are cleanly cropped and are more or less free from artifacts. Even the recently released FUNSD dataset of noisy scanned documents \cite{jaume2019} segment their words free of underlines, boxes, and spurious strokes. Consequently, most results on text recognition have reported their performance on clean test examples \cite{graves2009offline, bluche2016joint}, typically in the form of well-aligned, well-spaced text lines, which are not representative of the noisy, marked-up, richly formatted scanned documents encountered in the wild.

One possible way to improve the robustness of a text recognition system is to train it on images containing the types of artifacts typically present in documents, making it robust against such perturbations, a method akin to data augmentation or adversarial training \cite{goodfellow2015explaining}. However, today most organizations are already set up with industrial-grade recognition systems wrapped in cloud and security infrastructure, rendering the prospect of overhauling the existing system with a homemade classifier (which is likely trained on much fewer data and therefore a comparatively lower performance) too risky an endeavor for most.

Nonetheless, many industrial-grade classifiers are still not robust to document images with ink artifacts (Figure \ref{fig:cover}. An alternative way to address this problem is to \textit{erase} artifacts from the image before feeding it into the recognition engine. One might want artifact-cleansed images for other downstream tasks as well besides recognition, including signature extraction/verification \cite{signature2019} and document restoration, or simply for visual appeal; thus it is important to have an image pre-processing step that erases these artifacts.

Little work has been done leveraging deep learning for document artifact removal. In this work, we present DeepErase, which inputs a document text image with ink artifacts and outputs the same image with artifacts erased (Figure \ref{fig:cover}). Training is weakly supervised as we use a simple artifact assembler program to produce dirty images along with their segmentation masks for training. Note that henceforth we may refer to images with artifacts as ``dirty''. We evaluate the performance of DeepErase by passing the cleansed images into two popular text recognition tools: Tesseract and \texttt{SimpleHTR}. On these recognition engines, DeepErase achieves a 40-60\% word accuracy improvement (over the dirty images) on our validation set and a 14\% improvement on the NIST SDB2 and SDB6 datasets of scanned IRS documents.

\subsection{Related work}

Our work is related broadly to the field of semantic segmentation \cite{long2015fully, ronneberger2015u, badrinarayanan2017segnet}, which predicts classes for different regions of the image. While semantic segmentation is typically applied to natural scenes, several works have applied it to documents for page segmentation \cite{chen2017convolutional}, structure segmentation \cite{yang2017learning}, or text line segmentation \cite{renton2017handwritten}. All of these tasks discriminate large-scale structure within a document, such as tables or text lines, rather than small-scale patterns such as underlines striking through text characters.

Classical methods for line artifact detection used the Hough transform to detect lines and other simple shapes in documents, such as ellipses \cite{likforman1995hough, matas2000robust}. Such methods, however, do not pay attention to the spatial structure beyond specified shapes, and may erase parts of the clean text that overlapped with the artifact. Since the dawn of deep learning, similar tasks involving semantic segmentation in documents have been actively researched. Document binarization is a task in which each pixel in an RGB or grayscale image is assigned a binary value of either on or off. Binarization in low-contrast, degraded documents cannot rely solely on neighborhood-independent pixel thresholds and, like our task, must pay attention to the spatial patterns in the image. Recent approaches in binarization leverage multiscale convolutional networks to perform per-pixel binarization prediction  \cite{tensmeyer2017binarization}.

The works of Calvo-Zaragoza et al. \cite{calvo2017one} and K\"olsch et al. \cite{kolsch2018recognizing} are the most similar works to ours. The task in \cite{calvo2017one} is to discriminate between staff-lines and musical symbols in musical scores, while the task in \cite{kolsch2018recognizing} is to identify handwritten annotations inside of historical documents. LIke ours, both approaches leverage fully convolutional architectures for their respective semantic segmentation tasks. There are several differences which make our task more challenging. In \cite{calvo2017one}, the staff-lines and musical symbols, which the task wishes to distinguish, comprise a limited set of variations. Staff-lines appear in the same position with respect to the musical notes and tend to be long continuous horizontal lines. In contrast, our artifacts include lines, smudges, and spurious strokes in a variety of orientations and positions relative to the text. The historical document text characters in \cite{kolsch2018recognizing} are printed while the annotations are handwritten, and the annotations have a slightly different shade, both of which are telltale signs for the network to discriminate. Our images on the other hand are binarized before entering the model, forcing our segmentation to rely solely on neighborhood spatial structure. Finally, both these approaches require full supervision via manually labeled segmentation masks, while our approach is weakly supervised---only a single artifact image assembly function needs to be written.

\subsection{Contributions}
Our contributions are threefold:
\begin{itemize}
    \item \textbf{Novel application}: We tackle artifact removal in printed and handwritten text images, a problem not yet approached by deep learning. 
    \item \textbf{Weakly supervised approach}: Our approach requires only a clean, unlabeled set of printed or handwritten text images and artifacts which are widely available and a simple program to assemble them together. No manual pixel-level annotation is necessary.
    \item \textbf{Empirical results}: Our artifact-cleansed images achieve low test error and consequently have convincing performance upon visual inspection. Further, our artifact-cleansed images improve recognition accuracy on well-known text recognition engines such as Tesseract 4.0.
\end{itemize}

\section{Method}

\begin{figure*}[b!]
    \centering
    \begin{subfigure}[t]{.39\textwidth}
        \centering
        \includegraphics[width=.65\textwidth, trim=0cm 0cm 0cm 0cm, clip]{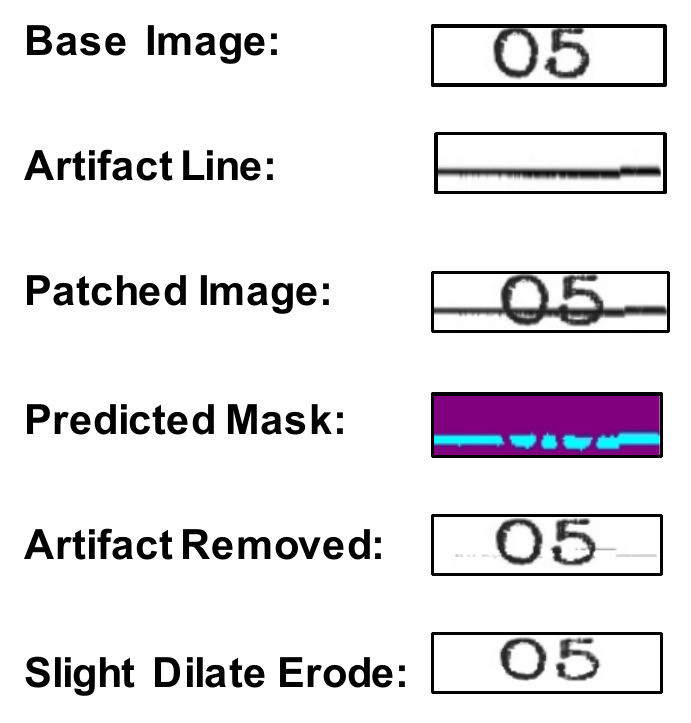}
        \caption{\small An artifact-patched image is obtained through the pixel-wise union of the base and artifact images. DeepErase predicts the artifact mask, which is used to remove artifacts.}
        \label{fig:surfaceflat-2}
    \end{subfigure}
    \quad\quad
    \begin{subfigure}[t]{.39\textwidth}
        \centering
        \includegraphics[width=.90\textwidth, trim=0cm 0cm 0cm 0cm, clip]{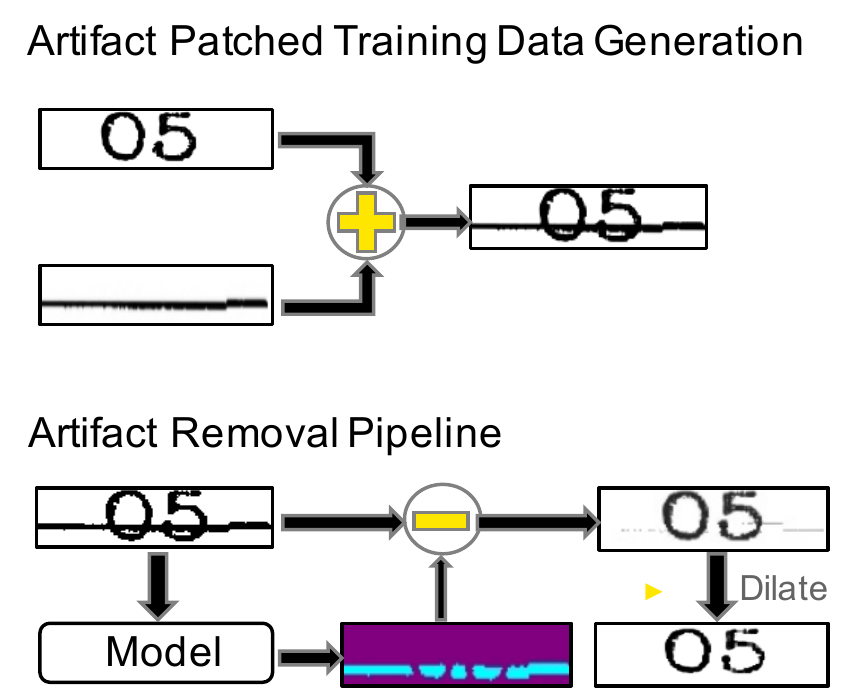}
        \caption{\small Flow diagrams showing the artifact patching and artifact removal procedures}
        \label{fig:procedure}
    \end{subfigure}
    \caption{Illustrations of how artifact text images are assembled for training, and how artifacts are removed during inference}
    \label{fig:my_label}
\end{figure*}

\subsection{Summary of approach}
Like other document binarization or segmentation tasks, we use a fully convolutional network to map the raw input image to a binary segmentation mask indicating artifact or no-artifact for each pixel in the image. Once the mask is obtained, all pixels on the mask indicating the presence of an artifact are set to 255 (white) on the input image, effectively cleansing it from artifacts. For training data, we automatically assemble a corpus of dirty images paired with their segmentation masks, generated using method described below in Section \ref{sec:genseg}, for both printed and handwritten text. The network is trained and validated on this data, and then tested in-the-wild on the NIST dataset of scanned IRS tax returns. Code for experiments is available at \url{https://github.com/yikeqicn/DeepErase}.

\subsection{Datasets}
In this work we train and test on both printed and handwritten text. Since printed text is easy to generate, we generate 280k text images in various fonts of words pulled from Wikipedia using \texttt{TextRecognitionDataGenerator} \cite{belval2019}. For handwritten text, we use the IAM dataset \cite{marti2002iam} consisting of about 110k handwritten words from 657 writers. For testing, we use the NIST SDB2 \cite{johnson2019nist} and NIST SDB6 \cite{johnson2019nist} datasets consisting of about 6k pages (each) of IRS tax return forms with printed and handwritten entries, respectively, each containing the types of artifacts that we wish to tackle in this work. We pre-crop text regions from the IRS dataset using image registration (the IRS documents all share the same template, making image registration especially effective) and manually defined crop regions for the template. In total, we have 22165 printed text images and 35202 handwritten text images from the IRS forms for testing. All images are binarized prior to being input into the model.

\subsection{Programmatic assembly of text images with artifacts}
\label{sec:genseg}
In order to automatically obtain a corpus of dirty images, we create a program which imposes realistic-looking artifacts on the readily available datasets of clean images. Similar ways of programmatically generating labeled data has been done for natural language processing tasks \cite{ratner2016data}. We focus on four types of artifacts: machine-printed underlines, machine-printed fill-in-the-blank boxes, random smudges, and handwritten spurious strokes. 

\begin{algorithm}[b]
\caption{Generation of text images with artifacts}
\label{alg:artifact}
\begin{algorithmic}[1]
    \STATE {\bfseries Input} clean image $\textbf{x}\in[0,255]^{n\times m}$, artifact sample $\textbf{x}_{art}\in[0,255]^{o\times p}$, offset
    \STATE {\bfseries Begin}
    \STATE \quad Binarize $\textbf{x}$ and $\textbf{x}_{art}$ with threshold of 128
    \STATE \quad Translate $\textbf{x}_{art}$ by offset, expanding image if needed \\
    \quad and filling additional pixels with intensity 255
    \STATE \quad Crop $\textbf{x}_{art}$ to the same size as $\textbf{x}$
    \STATE \quad Superimpose $\textbf{x}_{art}$ onto $\textbf{x}$ to get the dirty image, \\
    \quad i.e. $\textbf{x}_{dirty} \leftarrow \texttt{min(}\textbf{x}, \textbf{x}_{art}\texttt{)}$
    \STATE \quad Create segmentation mask, \\ \quad i.e. $\textbf{s}\leftarrow\textbf{x}_{art}+(255-\texttt{max(}\textbf{x}, \textbf{x}_{art}\texttt{)})$
    \STATE {\bfseries Return} dirty image $\textbf{x}_{dirty}$, segmentation mask $\textbf{s}$
\end{algorithmic}
\end{algorithm}

For random smudges and spurious strokes, we take a sampling of the IAM handwriting dataset to act as the artifacts. For line and box artifacts, we extract 5000 crops of horizontal and vertical lines and blank boxes from various sources of scanned forms, including the NIST IRS dataset as well as some internally scanned forms. See Figure \ref{fig:surfaceflat-2} for an example of a base image and an artifact used in the assembly process.

The datasets contain many examples of forms from the same template (e.g. the 1040 tax form). To automate extraction of lines or boxes, we first apply conventional homography-based image registration to the entire dataset, and then iteratively crop the same region from each image.

We then binarize both the clean and artifact images. This ensures that our network cannot rely on subtle differences in shading to predict artifacts. 

Next we sample an offset by which to translate the artifact image with respect to the clean image. This offset is sampled from a uniform distribution with bounds set such that the artifact falls within regions of the text that are consistent with the real-world. For instance, spurious strokes usually occur at the top or bottom of the image, while underlines usually occur at the bottom. We leave the boundaries of the distribution loose enough such that there is significant randomness and the artifacts overlap with the text characters a significant portion of the time.

After translating the artifact image by the offset amount, we then superimpose it onto the clean images by taking the lower intensity pixel (0 intensity corresponds to black) of the two (artifact and clean) images for each pixel in the clean image. Examples of the resulting dirty images are shown in Figure \ref{fig:valid}. The entire artifact text image generation pipeline is presented in Algorithm \ref{alg:artifact}. Figures \ref{fig:surfaceflat-2} shows examples of the intermediate images or masks and Figure \ref{fig:procedure} shows the artifact assembly (used during training) and removal (used during inference) pipelines.

Finally, the segmentation mask should contain all the markings of the artifact image minus the markings of the clean image. In other words, suppose that $A$ was the set of pixels containing the artifact marks, and $B$ is the set of pixels containing the clean marks. Then the segmentation mask (or pixels containing an artifact) would be $S = A - A\cap B$. During inference, once a segmentation mask is predicted, one can use it as a mask to erase the artifacts out of the image, as depicted in Figure \ref{fig:procedure}.

\subsection{Model architecture and training}
\begin{figure}[t]
\begin{center}
  \includegraphics[width=0.8\linewidth]{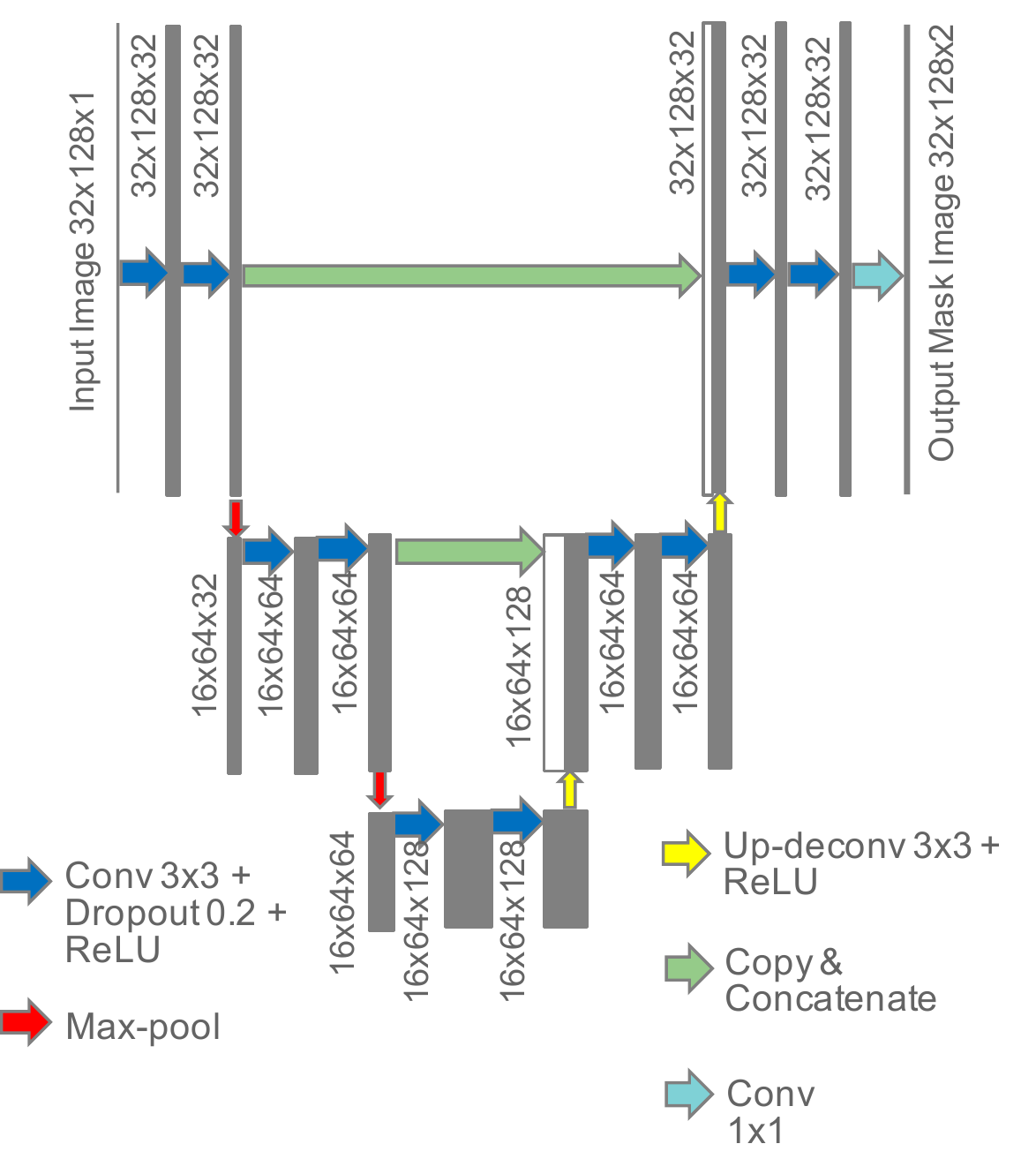}
\end{center}
\caption{Architecture for artifact segmentation}
\label{fig:arch}
\end{figure}

The network, schematic in Figure \ref{fig:arch} is a simple U-net architecture \cite{badrinarayanan2017segnet} which predicts a segmentation mask of artifact or no-artifact for each pixel. Convolutions are performed in blocks of two layers. At the end of each block, the feature map is downsampled via maxpooling, and the number of channels is doubled. After two blocks, the feature maps are upsampled via deconvolution (or transposed convolution) for two blocks until the feature map resolution is same as the original image. The first feature map in each upsampling block is concatenated with the last feature map from the corresponding downsampling block, as is done in U-net.

The training objective is simply to minimize the cross entropy loss between the true segmentation mask and the predicted segmentation mask on a per pixel basis, with averaging in the end. To address the class imbalance issue (there are a lot more pixels labeled not-artifact than as artifact) we use the median frequency balancing scheme from \cite{eigen2015predicting}. No regularizers are used in the training objective. The RMSProp optimizer is used to minimize the objective.

To encourage translation and size invariance, we apply data augmentation in the form of resizing, followed by horizontal and vertical shifts of the image within the fixed 32$\times$128 canvas.

\section{Evaluation}
\subsection{Comparative artifact detectors}
\label{sec:comp}
We compare DeepErase to two comparative artifact detectors.

\textbf{Hough}: The first is the widely used Hough-transform line detector, a classical computer vision method ubiquitous over the past several decades to detect and remove lines and other simple shapes from images. We utilize the standard OpenCV 3.0 Hough Line \cite{opencv2019} implementation.

\textbf{Manual Supervision CNN}: Second, we implement the approaches of \cite{calvo2017one} and also of \cite{kolsch2018recognizing} without ImageNet pretraining, which are nearly identical to ours except for the use of full, manual supervision. The authors of \cite{calvo2017one} manually annotated 20 scans of music documents for staff line removal. To be comparable, we manually annotated 60 document text images at the pixel level for training, costing about 3 man-hours. With such few examples, it is unlikely that the trained network will be able to model all the intricacies of artifact text, as we will see in Sections \ref{sec:val} and \ref{sec:irs}; this further highlights the need for weakly supervised approaches in order to achieve the dataset sizes needed for high model performance. We henceforth call this approach the ``Manual Supervision'' approach.

In our validation set results (Table \ref{tab:val}) we evaluate the Hough, Manual Supervision, and DeepErase approaches on a split of the datasets containing only line artifacts in order to ensure a fair comparison. Since the error for Manual Supervision and DeepErase on the line-artifacts-only split was always lower than its error for the entire dataset, we report only the error on the entire dataset for Manual Supervision and DeepErase.

Since the Hough approach is validated on a split of the full validation set, it has a different value for recognition accuracy on dirty images in Table \ref{tab:val}. Meanwhile the IRS dataset is consisted entirely of line (vertical or horizontal) artifacts so the dirty recognition accuracies in Table \ref{tab:irs} are identical.

\subsection{Metrics}
Other than visual inspection, we use two metrics to determine our performance on artifact removal.

\textbf{Segmentation error}: First, we use the segmentation error on the validation set, which is the probability that a pixel on the predicted segmentation mask does not match the ground truth. \textit{Baseline}: to compare our results, we include the segmentation error on the original clean text images before artifact assembly, which has a ground-truth segmentation mask that is uniformly annotated with no-artifact. This baseline ensures that when the artifact detector sees an image with no artfact inside, it does not falsely claim that there are artifacts.

\begin{table}[t]
\centering
\caption{Segmentation results on validation set}
\begin{tabular}{l|cc}
\hline
                         & \multicolumn{2}{c}{Segmentation error}  \\
            Setting             & Baseline & Cleaned                              \\ 
\hline
Hough on printed         & 1.2      &   17.62                                  \\
Manual Supervision on printed & 0.55 & 6.16 \\
DeepErase on printed    & 0.4      & \textbf{3.38}                                   \\
\hline
Hough on handwritten         & 0.56      & 15.31  \\
Manual Supervision on handwritten & 0.45 & 7.20 \\
DeepErase on handwritten    & 0.25      & \textbf{4.36}                                   \\
\hline
\end{tabular}
\label{tab:seg}
\end{table}

\textbf{Recognition error}: The secondary metric that we use for evaluating performance is recognition error. The simple assumption is that images cleaned from artifacts will make it easier for recognition models to discriminate. Two recognition error metrics are reported. Character error rate (CER) is the string edit distance between the predicted string and the ground truth string, or in other words, the minimum number of per-character add, delete, or replace operations needed to match the two strings. Word error rate (WER) is the probability that the predicted word does not match the ground truth, regardless of how far off it is. \textit{Baseline}: Like the baseline for segmentation error, we use the recognition accuracy on the ``gold-standard'' original clean images without any artifacts superimposed as our recognition baseline. These are the raw unmodified images from \texttt{TextRecognitionDataGeneratoror} for printed and IAM for handwritten.

For printed text recognition we use the widely used open-source Tesseract v4 software. Since there is no widely available offline handwriting recognition software, we used the model from the \texttt{SimpleHTR} repo \cite{simple2019}. Both softwares are based on an LSTM-CTC architecture.

\begin{figure*}
    \begin{center}
      \includegraphics[width=0.99\linewidth]{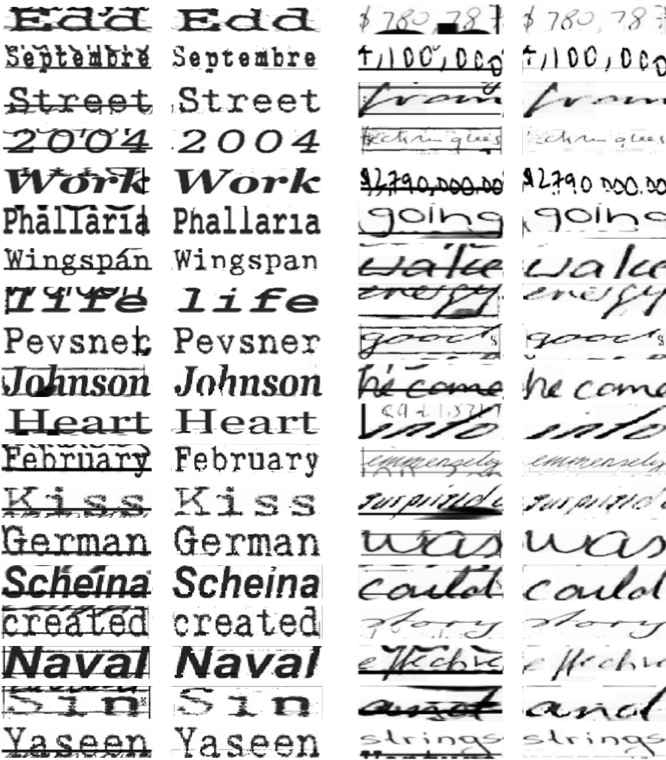}
    \end{center}
   \caption{Examples from validation results. Columns 1 and 3 are before cleansing, 2 and 4 are after cleansing.}
    \label{fig:valid}
\end{figure*}

\begin{table*}
\centering
\caption{Recognition results on validation sets}
\begin{tabular}{l|cc|cc|cc} 
\hline
                                               & \multicolumn{2}{c|}{Baseline} & \multicolumn{2}{c|}{Dirty} & \multicolumn{2}{c}{Cleaned}  \\
Setting                                        & CER & WER                 & CER & WER                  & CER                   & WER  \\ 
\hline
Hough on printed         &  13.23   & 20.89          & 129.53 & 95.05             & 132.83                & 93.67 \\
Manual Supervision on printed & 13.23 & 20.89 & 104.98 & 93.89 & 53.12 & 54.94 \\
DeepErase on printed            & 13.23    & 20.89          & 104.98 & 93.89             & 29.13                 & \textbf{34.71} \\ 
\hline
Hough on handwritten     & 6.89    & 20.22         &  50.51   & 78.34 & 52.32 & 81.71     \\
Manual Supervision on handwritten & 6.89 & 20.22 & 46.24 & 77.78 & 37.63 & 66.67 \\
DeepErase on handwritten       & 6.89   & 20.22 &    46.24 & 77.78       & 28.58 & \textbf{47.20}     \\
\hline
\end{tabular}
\label{tab:val}
\end{table*}

\begin{table*}
\centering
\caption{Recognition results on NIST IRS datasets}
\begin{tabular}{l|cc|cc} 
\hline
                         & \multicolumn{2}{c|}{Baseline} & \multicolumn{2}{c}{Cleaned}  \\ 
Setting                  & CER & WER                 & CER & WER                    \\ 
\hline
Hough on printed         &  97.26 & 78.87 & 194.13    & 94.98                       \\
Manual Supervision on printed & 97.26 & 78.87 & 67.66 & 73.89 \\
DeepErase on printed     &  97.26 & 78.87 & 60.87    & \textbf{64.20}                       \\
\hline
Hough on handwritten     & 94.93    & 98.38  & 81.19      & 93.09                       \\
Manual Supervision on handwritten & 94.93 & 98.38 & 70.04 & 91.18 \\
DeepErase on handwritten & 94.93    & 98.38 & 59.91    & \textbf{84.86}                       \\
\hline
\end{tabular}
\label{tab:irs}
\end{table*}

\subsection{Validation results}
\label{sec:val}
We first test our model on a held-out set of examples from our dirty datasets. Since we used a train/validation split of 9:1, the held-out set consists of 28k examples for printed and about 11k for handwritten. Since our dirty dataset was crafted from a base dataset (raw images from \texttt{TextRecognitionDataGenerator} or IAM), we report the performance of the original base images (which do not have artifacts) on the recognition models as our baseline.

Using DeepErase, we observe segmentation error of less than 5\% on printed and handwritten text, which means that most pixels are correctly erased (see Table \ref{tab:seg}). In contrast, the Hough transform-based line removal achieves significantly higher error, since it removes entire lines including the parts which overlap with the text. The Manual Supervision approach performs better than Hough, but does not achieve as low of error as DeepErase, due to the shortage of available Manual Supervision data as discussed in Sec. \ref{sec:comp}.

Good segmentation leads to greatly improved recognition performance as well as shown in Table \ref{tab:val}. When the artifacts are erased before inputting into Tesseract or \texttt{SimpleHTR}, the recognition accuracy improves by 60.56\% and 31.20\%, respectively, compared to no cleaning. DeepErase-cleaned images also achieve 20-60\% lower downstream recognition word error than those clean by the Hough and Manual Supervision approaches. The segmentation is not perfect though---when compared with the ``gold standard'' base images, cleansed images get about 15-30\% higher recognition error. Figure \ref{fig:valid} shows some example images before and after artifact erasing.

\subsection{Results on real-world NIST IRS dataset}
\label{sec:irs}
In addition to evaluating on the validation set, we wish to test DeepErase in the wild on text from scanned IRS tax return forms. In-the-wild data tends to experience distribution shift \cite{quionero2009dataset}, leading to lower performance when tested on models trained on data from other distributions. Typically this results in an iterative process where the training data is better adapted to the distribution in-the-wild, and the system is re-tested. We present results from our first-pass here, where we had not seen the IRS data before designing our artifact generation algorithm \ref{alg:artifact}.

On the IRS printed data, removing artifacts via DeepErase lowers the Tesseract recognition error by 14.67\% compared to not removing them, as shown in Table \ref{tab:irs}. Similarly on the handwritten data, it lowers the \texttt{SimpleHTR} recognition error by 13.52\%. In both cases, DeepErase performs better than the Hough and Manual Supervision comparables.

\begin{figure*}[t]
    \begin{center}
      \includegraphics[width=0.99\linewidth]{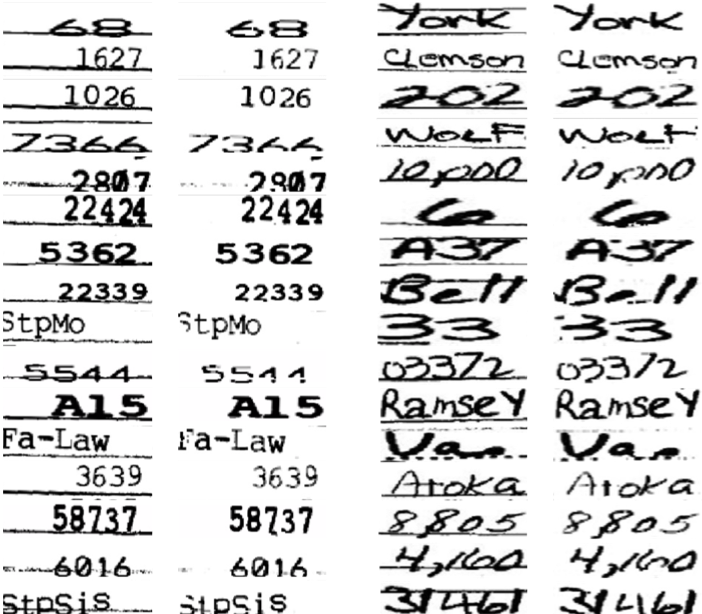}
    \end{center}
   \caption{Examples from IRS results. Columns 1 and 3 are before cleansing, 2 and 4 are after cleansing.}
    \label{fig:test}
\end{figure*}

Figure \ref{fig:test} shows examples of artifact removal in both printed and handwritten IRS text. Despite the relatively high recognition error on handwritten data even after cleaning (which is primarily due to distribution shift), upon visual inspection the erased images look reasonably good and indicate that the objective of artifact removal (to yield better results on other downstream recognition engines or other tasks) is satisfied.

\section{Conclusion}

We have presented DeepErase, a neural-based approach to removing artifacts from document text images. This task is challenging because it must rely solely on spatial structure (rather than differences in shading since the images are binarized) to do semantic segmentation of a wide variety of artifacts. We present a method to programmatically assemble unlimited realistic-looking text artifact images from real data and use them to train DeepErase in weakly supervised manner. The results on the validation set are excellent, showing good segmentation along with a 40 to 60\% boost in recognition accuracy for both printed and handwritten text using common recognition software. On the real-world IRS dataset, DeepErase improves recognition accuracy by about 14\% on both printed and handwritten text. The cleansed images on both printed and handwritten examples look visually convincing. Next steps include better modeling the test distribution during the artifact generation process such that the trained model performs better at test time.

\newpage

\section*{Acknowledgements}
The authors thank Darrin Williams, Sameer Gupta, Carl Case, Ali Khan, and Rajarajan Sampath for project oversight and support. The results discussed in this
letter and references to terms architecture, robustness, efficient, accurate, and bias are with respect to the letter’s mathematical treatment of a generalized methodology framework. All data used in this work were from the public domain, and the text images therein were selected at random and should be be interpreted to represent any more than examples for the algorithm demonstration. The views and conclusions expressed in this material are those of the authors and should not be interpreted as representing the official policies or endorsements, either expressed or implied, of Ernst \& Young LLP. Software from \url{https://comet.ml} accelerated this work.

{\small
\bibliographystyle{ieee}
\bibliography{egbib}
}

\appendix

\input{supp.tex}

\end{document}

%% file: supp.tex

\begin{figure*}[hb!]
    \begin{center}
      \includegraphics[width=0.7\linewidth]{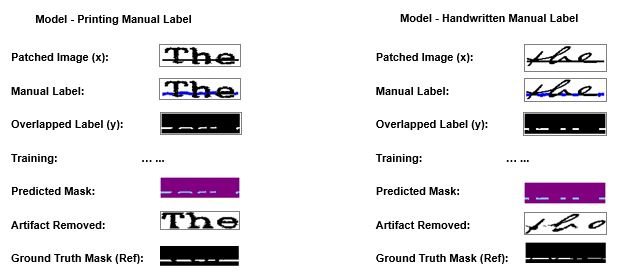}
    \end{center}
   \caption{Manual Supervision CNN Model Training and Usage}
    \label{fig:manual_model}
\end{figure*}

\section{Manual Supervision CNN Model}

We have provided a set of illustrations around the DeepErase model. In the appendix, we are presenting further details around the benchmark "Manual Supervision CNN" model (Section 3.1 in the main paper).

In Section 3.1 main paper, we described the training method for the benchmark Manual Supervision CNN Model. In Figure \ref{fig:manual_model}, we presents the process of model training as below:
\begin{itemize}
    \item \textbf{Training Images}: The 60 synthetic images with artifacts were used as development images. See "Patched Image (x)" in the Figure \ref{fig:manual_model}. 
    \item \textbf{Manual Coloring}: The artifact areas in the images were colored manually. See "Manual Label" in Figure \ref{fig:manual_model}.
    \item \textbf{Finalize Labeling}: The manual colored pixels were compared against the dark areas of original patched images. The pixels of the overlapped areas were verified as positive pixels. See "Overlapped Label(y)" in Figure \ref{fig:manual_model}.
\end{itemize}
The model prediction is the same as the DeepErase Model, which includes the following procedures:
\begin{itemize}
    \item \textbf{Predict Masks}: Apply the model upon images with artifacts. The pixels of artifact areas would be predicted as positive. See "Predicted Mask" in the Figure \ref{fig:manual_model}. 
    \item \textbf{Remove Artifacts}: Force the identified artifact area to be the background color. See "Artifact Removal" in Figure \ref{fig:manual_model}.
    \item \textbf{Downstream Process}: The cleaned images would be ready to use for downstream process. i.e. text recognition.
\end{itemize}
\section{Text Recognition Result Examples}
\begin{figure*}[t!]
    \begin{center}
      \includegraphics[width=0.7\linewidth]{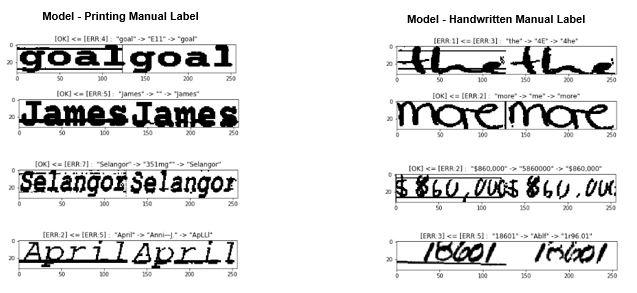}
    \end{center}
   \caption{Manual Supervision CNN Model - Text Recognition Results - Validation Datasets}
    \label{fig:man_model_rvalidation_results}
\end{figure*}

\begin{figure*}
    \begin{center}
      \includegraphics[width=0.7\linewidth]{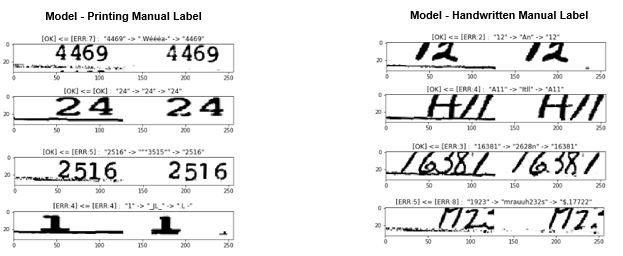}
    \end{center}
   \caption{Manual Supervision CNN Model - Text Recognition Results - IRS NIST Datasets}
    \label{fig:man_model_rirs_results}
\end{figure*}

In Section 3.3 and 3.4 main paper, we discussed the text recognition results for validation data and real-world NIST IRS datasets. In this section, we attach a few of result images as illustration. Overall, similar to DeepErase model, the Manual Supervision model was able to help remove artifacts.However, The removal was less accurate due to limited labeled training data and potentially less accurate manual labelling. The removal could be incomplete or overactive. See the result images in Figure \ref{fig:man_model_rvalidation_results} and Figure \ref{fig:man_model_rirs_results}.

